\documentclass[preprint,12pt]{elsarticle}

\usepackage{amssymb}

\usepackage{graphicx} %package to manage images
\graphicspath{ {./images/} }

\usepackage[rightcaption]{sidecap}
\usepackage{wrapfig}
\usepackage{hyperref}
\usepackage{caption}
\usepackage{subcaption}
\usepackage{float}

\usepackage[T1]{fontenc}

\usepackage[ruled,vlined]{algorithm2e}
\usepackage{tabularray}

\journal{Pattern Recognition Letters}

\begin{document}

\begin{frontmatter}

\title{Sparse then Prune: Toward Efficient Vision Transformers}

\author[inst1]{Yogi Prasetyo}

\affiliation[inst1]{organization={Informatics Department, Faculty of Computer Science, Universitas Brawijaya},%Department and Organization
            addressline={Jalan Veteran 8}, 
            city={Malang},
            postcode={65145}, 
            state={Malang},
            country={Indonesia}}

\author[inst1]{Novanto Yudistira}
\author[inst1]{Agus Wahyu Widodo}

\begin{abstract}
%% Text of abstract
 
The Vision Transformer architecture is a deep learning model inspired by the success of the Transformer model in Natural Language Processing. However, the self-attention mechanism, large number of parameters, and the requirement for a substantial amount of training data still make Vision Transformers computationally burdensome. In this research, we investigate the possibility of applying Sparse Regularization to Vision Transformers and the impact of Pruning, either after Sparse Regularization or without it, on the trade-off between performance and efficiency. To accomplish this, we apply Sparse Regularization and Pruning methods to the Vision Transformer architecture for image classification tasks on the CIFAR-10, CIFAR-100, and ImageNet-100 datasets. The training process for the Vision Transformer model consists of two parts: pre-training and fine-tuning. Pre-training utilizes ImageNet21K data, followed by fine-tuning for 20 epochs. The results show that when testing with CIFAR-100 and ImageNet-100 data, models with Sparse Regularization can increase accuracy by 0.12\%. Furthermore, applying pruning to models with Sparse Regularization yields even better results. Specifically, it increases the average accuracy by 0.568\% on CIFAR-10 data, 1.764\% on CIFAR-100, and 0.256\% on ImageNet-100 data compared to pruning models without Sparse Regularization. Code can be accesed here: \href{https://github.com/yogiprsty/Sparse-ViT}{https://github.com/yogiprsty/Sparse-ViT}
\end{abstract}


\end{frontmatter}

%% \linenumbers

%% main text
\pagebreak

\section{Introduction}
\label{sec:introduction}
Modern information technology advancements make it simpler and faster for people to tackle a variety of difficulties. Humans produce a lot of data daily due to its broad usage, including search engine activity, online photo and video uploads, and more. One effective approach to automate data classification is by utilizing Deep Learning technology. Recently, the most popular deep learning architecture in digital image processing is the Convolutional Neural Network (CNN) due to its success in the ILSVRC 2014 GoogleNet challenge\cite{DBLP:journals/corr/SzegedyLJSRAEVR14}. However, this architecture generally requires a relatively high computational time, making it possible for new research to develop an architecture that can outperform CNN.

The self-attention-based architecture, especially the Transformer, has become the preferred choice in natural language processing. In some cases, the Transformer significantly outperforms convolutional and recurrent layer-based architectures\cite{DBLP:journals/corr/VaswaniSPUJGKP17}. Not only in natural language processing, but the use of Transformers is also starting to expand, including in digital image processing. To apply the Vision Transformer, start by dividing the image into patches or small pieces and provide the sequence of positions of these patches as input to the Transformer layer. Patches of images are treated like tokens in natural language processing applications. This method is five times faster than EfficientNet, which is based on a Convolutional Neural Network. The accuracy of the Vision Transformer model will increase significantly if it is pre-trained with large data, such as ImageNet21k, before using it for simpler cases\cite{DBLP:journals/corr/abs-2010-11929}.

In addition to researching better architectures, researchers have proposed various methods to improve accuracy or speed up computing in Deep Learning models. The widely used methods include Regularization and Pruning. Regularization is an additional technique that aims to enhance the model's generalization, resulting in better results on test data\cite{DBLP:journals/corr/abs-1710-10686}. On the other hand, pruning is a popular technique used to obtain a more compact model by removing weights considered less important\cite{DBLP:journals/corr/abs-2103-03014}. This approach allows for faster computation time while maintaining the model's accuracy.

Based on the above background, this research applies Sparse Regularization and Pruning methods to the Vision Transformer architecture for image classification tasks utilizing the CIFAR 10 and CIFAR 100 datasets. The CIFAR 10 dataset comprises 60,000 images with a resolution of 32x32, divided into ten distinct classes, each consisting of 6,000 images. In contrast, the CIFAR 100 dataset includes 60,000 images of the same resolution but encompasses a broader spectrum of classes, specifically 100 classes. As an open-source dataset extensively utilized in Deep Learning research, CIFAR datasets serve as an ideal testbed for evaluating the proposed techniques and their impact on image classification tasks.

\section{Related Work}
\label{sec:related-works}
Research conducted by \cite{DBLP:journals/corr/abs-2010-11929}, titled "An Image is Worth 16x16 Words: Transformers for Image Recognition at Scale," attempts to apply pre-training to the Vision Transformer architecture using varying amounts of data. Consequently, when trained on small data, the Vision Transformer model exhibits poor performance. However, when trained on large data, this model demonstrates superior performance and surpasses the ResNet model. Other research also shows the same conclusion when the Vision Transformer is used to classify mask usage using the MaskedFace-Net dataset. As a result, the pre-trained ViT-H/14 model performed better than the ViT-H/14 model without pre-trained, with 1.87\% higher on test accuracy\cite{JAHJA2023200186}. The findings of this research have inspired the utilization of pre-trained models in the present study.

Much research has been done to improve Vision Transformer-based architecture. The Data-efficient image Transformers (DeiT) \cite{DBLP:journals/corr/abs-2012-12877} introduce a CNN model as a teacher and apply knowledge distillation\cite{hinton2015distilling} to enhance the student model of ViT in order to lower its dependence on a huge amount of data. To obtain satisfactory results thus, DeiT only can be trained on ImageNet. Additionally, the CNN models and distillation types selected may affect the final performance. The Convolution enhanced Image Transformer (CeiT)\cite{yuan2021incorporating} proposed to overcome those limitations. This method works by combining the advantages of CNN in low-feature extraction, strengthening the locality of the advantages of Transformers in establishing long-range dependencies. The Convolutional Vision Transformer (CvT)\cite{DBLP:journals/corr/abs-2103-15808} proposed to improve the performance and efficiency of ViTs by introducing convolutions into the ViT architecture. This design introduces convolutions in two core sections of ViT: a hierarchy of Transformers containing a new convolutional token embedding and a convolutional Transformer block utilizing a convolutional projection.

Another method called sharpness-aware minimizer (SAM) was proposed and utilized to explicitly smooths the loss geometry during the training process\cite{DBLP:journals/corr/abs-2106-01548}. SAM seeks to find solutions that the entire environment has low losses rather than focusing on any single point. The Transformer-iN-Transformer (TNT)\cite{DBLP:journals/corr/abs-2103-00112} was proposed to enhance the ability of feature representation in ViT by dividing the input images into several patches as "visual sentence" and then divide those patches into sub-patches, as the representation of "visual word". Apart from using conventional transformer blocks to extract features and attention of visual sentences, sub-transformers are also embedded into the architecture to excavate the features and details of smaller visual words.

Another research conducted by \cite{7966185}, titled "Improvement of Learning for CNN with ReLU Activation by Sparse Regularization," applies the Sparse Regularization method to ReLU inputs. The findings demonstrate a significant increase in accuracy, ranging from 9.98\% to 12.12\%, compared to models without Sparse Regularization. Additionally, compared to the Batch Normalization method, the accuracy improves by 4.64\% to 6.87\%. These compelling results have inspired using the Sparse Regularization method in the present study.

The research conducted by \cite{DBLP:journals/corr/abs-2104-08500}, titled "Vision Transformer Pruning," applies the Pruning method to the Vision Transformer architecture. The findings demonstrate that this method reduces FLOPS (Floating Point Operations per Second) by 55.5\% when applied to the Vision Transformer Base 16 architecture. Moreover, the accuracy only decreases by 2\% on ImageNet-1K data. These promising results have inspired using the Pruning method in the present study.

Previous studies have shown that Vision Transformer active neurons are highly sparse. This implies that only a small portion of neurons within a layer or network become active or "fire" at any given moment. In contrast, most neurons remain inactive or have low activation values\cite{DBLP:journals/corr/abs-2106-01548}. These findings indicate that less than 10\% of neurons in a Vision Transformer model have values greater than zero. In simpler terms, this highlights the significant potential of Vision Transformer models for network pruning.

\section{Method}
\label{sec:method}

\subsection{Datasets}
\label{datasets}
We utilized the widely recognized CIFAR-10 and CIFAR-100 datasets as standard benchmarks during the experiments. CIFAR-10 comprises RGB color images categorized into ten classes for object classification, while CIFAR-100 consists of 100 different classes. Each image in these datasets had been standardized to a resolution of 32x32 pixels. This consistency in size and format allowed us to focus solely on the intricacies of the data and the impact of various regularization methods on model performance. The standardized resolution facilitated seamless comparisons between different models and enabled us to discern subtle nuances that might have otherwise gone unnoticed. CIFAR datasets is commonly used in research for evaluating performance. The research conducted by \cite{DBLP:journals/corr/Gastaldi17} employs the CIFAR-10 and CIFAR-100 datasets to evaluate the effectiveness of this regularization method. By conducting experiments on CIFAR datasets, researchers can gauge the impact of regularization methods on model performance and compare them to existing approaches.

\subsection{Vision Transformer}
\label{vision-transformer}
\begin{figure}[h]
    \centering
    \includegraphics[width=\textwidth]{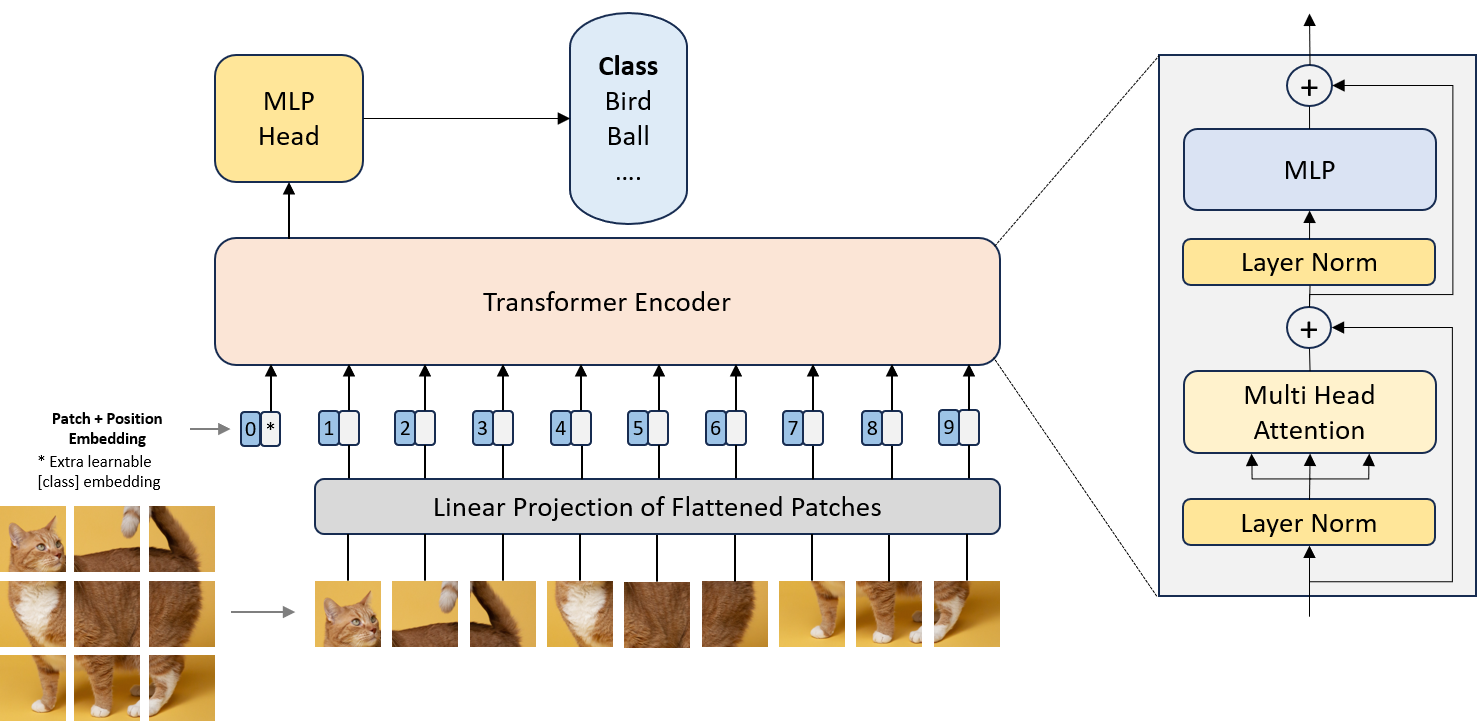}
    \caption{Vision Transformer\cite{DBLP:journals/corr/abs-2010-11929}}
    \label{fig:vision-transformer}
\end{figure}

The Transformer is an attention-based model published in 2017 to replace the Recurrent layer-based architecture, addressing issues with gradients and relatively long computation time\cite{DBLP:journals/corr/VaswaniSPUJGKP17}. The Vision Transformer is an architecture inspired by the success of the Transformer in natural language processing. In 2020, Transformers were employed in image classification problems and demonstrated superior performance compared to CNN-based architectures, such as EfficientNet\cite{DBLP:journals/corr/abs-2010-11929}. Figure \ref{fig:vision-transformer} illustrates the architecture of the Vision Transformer.

Unlike state-of-the-art architecture, The process begins by breaking the 2D images into several parts and converting them into a sequence of flattened 2D patches or 1-dimensional vector, shown in Equation \ref{eq:patch-emb}, where $(H,W)$ is the resolution of the original image, $C$ is the number of channels, $P$ is the number of patches and $N = HW | P^2$. During this process, position embedding is added to the vectors, and these vectors are combined with an additional vector (class token) that will serve as the output of the Vision Transformer architecture.

Images that have undergone the Patch and Embedding process will pass through Transformer Encoder, which includes the Multi-Head Self-Attention (MSA) layer and the Multi-Layer Perceptron (MLP). The normalization layer is applied before the MSA and the MLP. Layer normalization is used to reduce the computation time of the artificial neural network. This layer operates by calculating the average and variance of each data input, which are then used to normalize all inputs \cite{ba2016layer}. Equation \ref{eq:norm-eq} represents the formula for the normalization layer, where $x_{i,k}$ is vector input.

\begin{equation} \label{eq:patch-emb}
    x \in R^{HWC} \rightarrow x_p \in R^{N(P^2C)}
\end{equation}

\begin{equation} \label{eq:norm-eq}
    \hat{x}_{i,k} = \frac{x_{i,k} - \mu^i}{\sqrt{\sigma_i^2 + \epsilon}}
\end{equation}

The result will pass through Multi-Head Attention, and this layer works by dividing the input into several heads, allowing each head to learn a different level of Attention. The heads' outputs are combined and forwarded to the Multi-Layer Perceptron. Equation \ref{eq:attn} represents the formula for the Scaled Dot-Product Attention, while Equation \ref{eq:mha} represents the formula for the Multi-Head Attention.

\begin{equation} \label{eq:attn}
    head(Q,K,V) = softmax \left(\frac{QK^t}{\sqrt{d_k}} \right) V
\end{equation}

\begin{equation} \label{eq:mha}
    MultiHead(Q,K,V) = Concat(head_1\, ...\, head_n)
\end{equation}

The last layer of transformer encoding is the multilayer perceptron (MLP), an artificial neural network architecture with one or more hidden layers between the input and output layers. Training an MLP model involves adjusting the weights and biases between neurons at different layers to minimize the difference between the model's output and the desired output\cite{articleRamchoun}. The MLP layer used Gaussian Error Linear Unit (GELU). In certain datasets, these non-linear activation functions can match and outperform linear functions such as ReLU and ELU\cite{DBLP:journals/corr/HendrycksG16}. The equation for the GELU activation function can be seen in Equation \ref{eq:gelu}.

\begin{equation} \label{eq:gelu}
    GELU(x) = 0.5x\left(1+tanh\left[\sqrt{2/\pi}\left(x+0.044715x^3\right)\right]\right)
\end{equation}

\subsection{Sparse Regularization}
\label{sec:sparse-regularization}
Sparse regularization is a leading technique in deep learning research that aims to improve the performance and efficiency of models by promoting sparsity in the learned weights. This regularization method encourages a subset of the weights to become zero or close to zero, resulting in a more compact and interpretable model\cite{7966185}. Also, this method has a similar effect to Batch Normalization and can improve the model's ability to predict more general data. Evaluating sparseness can be done in various ways. However, in the research, sparse regularization is applied to the ReLU input using Equation \ref{eq:sparse}, where $h_k$ is the input the $k$-th ReLU.
\begin{equation} \label{eq:sparse}
    S(h_k) = log(1+h_k^2)
\end{equation}
Once the sparse value is obtained, it is multiplied by $\lambda$, a parameter that controls sparseness, before being added to the loss value $L$. Further details can be found in Equation \ref{eq:new-loss}.
\begin{equation} \label{eq:new-loss}
    E = L + \lambda\sum_{k}S(h_k)
\end{equation}

\begin{figure}[h]
     \centering
     \begin{subfigure}[b]{0.45\textwidth}
         \centering
         \includegraphics[width=\textwidth]{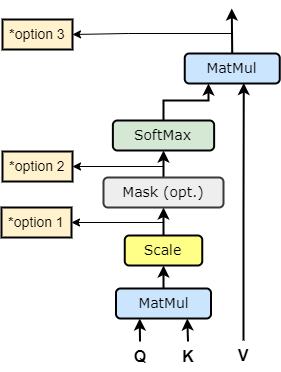}
         \caption{}
         \label{fig:sparse-mha}
     \end{subfigure}
     \hfill
     \begin{subfigure}[b]{0.45\textwidth}
         \centering
         \includegraphics[width=\textwidth]{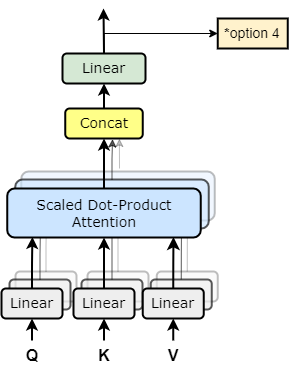}
         \caption{}
         \label{fig:sparse-attn}
     \end{subfigure}
     \caption{Choices to put sparse regularization}
\end{figure}

Many explorations have taken place in the field of regularization techniques within deep learning to determine optimal placements within models. Previous studies have provided insights into the effectiveness of sparse regularization in convolutional neural network (CNN) models, particularly when applied to the input of ReLU\cite{7966185}. Figure 2 demonstrates a range of interesting options for strategically incorporating sparse regularization within the model architecture, each offering unique insights and potential benefits.

Option 1 suggests an interesting placement for sparse regularization, where sparsity is evaluated after multiplying queries and keys, followed by scaling, also known as the similarity score. In option 2, sparsity is evaluated precisely at the attention weight stage. This value is obtained after the similarity score passes through the softmax activation function. Option 3 applied sparse regularization after multiplying attention weight with V (value), also known as weighted value. Option 4, sparsity, is evaluated after the linear or output layer in Multi-Head Attention. Last option, according to previous research, we will apply sparse regularization to the input of GELU activation function in MLP layer.

\subsection{Pruning}
\label{sec:pruning}
Pruning is a method used to speed up computation in deep learning models by removing the least important parameters. Various approaches for deleting unneeded parameters can be used, such as using the $l_1$-norm or $l_2$-norm to select the target parameters for deletion\cite{DBLP:journals/corr/abs-2103-03014}. Pruning is broadly classified into unstructured and structured pruning, each with its strategy for parameter reduction\cite{DBLP:journals/corr/abs-2007-00389}.

Structured pruning operates by systematically removing structured components within deep learning models. These components may manifest as filters\cite{DBLP:journals/corr/abs-1808-06866}, channels, or even entire layers, enabling a more holistic approach to parameter reduction\cite{he2023structured}. By selectively excising these structured elements, deep learning models can be streamlined, improving efficiency and reducing computational complexity.

In contrast, unstructured pruning (algorithm \ref{alg:prune}) focuses on individual weights within layers or filters. This surgical precision allows for the precise removal of individual weights without disrupting the overall arrangement of the deep learning model. Unstructured pruning can still achieve significant parameter reduction while maintaining the accuracy of the pruned network\cite{laurent2020revisiting}, resulting in faster computations and optimized model performance.

\begin{algorithm}[H]
    \SetKwInOut{Input}{Input}
    \SetKwInOut{Output}{Output}
    
    \Input{model, pruning\_ratio}
    \Output{model}
    
    parameters $\gets$ []\;
    \For{param \textbf{in} model.parameters()}{
        parameters.append(param)\;
    }
    all\_parameters $\gets$ torch.cat([param.data.view(-1) \textbf{for} param \textbf{in} parameters])\;
    threshold $\gets$ torch.kthvalue(torch.abs(all\_parameters), int(pruning\_ratio * all\_parameters.numel())).values\;
    \For{param \textbf{in} parameters}{
        param\_data $\gets$ param.data\;
        mask $\gets$ torch.abs(param\_data) \textgreater threshold\;
        pruned\_param $\gets$ param\_data * mask.float()\;
        param.data $\gets$ pruned\_param\;
    }
    \Return{model}\;
    
    \caption{Global Pruning with L1 Unstructured Method}
\label{alg:prune}
\end{algorithm}

\subsection{Transfer Learning}
\label{transfer-learning}
The Vision Transformer performs remarkably when complemented by a pre-training process known as transfer learning \cite{DBLP:journals/corr/abs-2010-11929}. Transfer learning involves training the model on a large dataset before applying it to other data, leveraging the knowledge acquired from the pre-training phase to enhance performance on new tasks. In this context, the abundance of data and its diverse nature play pivotal roles in facilitating effective transfer learning. Large amounts of data and a wide diversity are the most influential factors for transfer learning \cite{Weiss2016}. To exploit the full potential of transfer learning, this research endeavors to capitalize on the Vision Transformer architecture, which has undergone extensive training using ImageNet21k data. ImageNet21k boasts an impressive collection of 14,197,122 annotated images, accurately organized according to the WordNet hierarchy\cite{Russakovsky2015}. This vast and diverse dataset encapsulates numerous objects, scenes, and concepts, providing rich visual information for the Vision Transformer to learn from.

\subsection{Model Training}
\label{model-training}
The ViT architecture can be configured in multiple ways. To ensure unbiased test results for each configuration, constant hyperparameters will be set during training for the ViT architecture's various configurations. For this research, we will maintain the same setup used in the original ViT research, as described in reference \cite{DBLP:journals/corr/abs-2010-11929}. During the fine-tuning process, the following hyperparameters were employed: a batch size of 64, a learning rate of 0.03, 20 epochs, a Cross-Entropy loss function\cite{DBLP:journals/corr/abs-1805-07836}, Stochastic Gradient Descent optimizer, and GeLU activation function\cite{DBLP:journals/corr/HendrycksG16}. For transfer learning, we utilized pre-trained weights initially trained on the ImageNet21K dataset\cite{5206848}. The experiments were conducted using hardware specifications that included an RTX 8000 GPU, an Intel(R) Xeon(R) Gold 6230R processor, and 255 GiB of RAM.

\section{Experiment}
\label{experiemnt}
To confirm the effectiveness of the implemented method, we have performed experiments on the testing of the Vision Transformer using CIFAR-10 and CIFAR-100 datasets.

\subsection{ViT Architecture}
\label{vit-arch}
According to the original research paper, the authors introduced several Vision Transformer (ViT) model variants. They explored different model configurations to investigate the impact of architectural choices on its performance. There are three variants of the vision transformer model. The first is ViT-Base. This variant serves as the baseline model, featuring a transformer architecture with a moderate number of layers and attention heads. It serves as a reference point for comparing the performance of other ViT models. The Second is ViT-large. This variant extends the ViT model with more layers and attention heads, increasing its capacity and potential for capturing complex visual patterns. ViT-large aims to achieve higher accuracy by leveraging a deeper and more parameter-rich architecture. The last is ViT-Huge; as the name suggests, this variant represents an even more expansive and powerful instantiation of the Vision Transformer. It features significantly larger layers and attention heads, providing a massive capacity for learning intricate visual representations. We have done the experiments using ViT-B/16 architecture. More detailed configuration and hyperparameter for Vision Transformer are shown in the Tabel \ref{table:vit-arch}.

\begin{table}[h]
\centering
\caption{Configurations and Hyperparameters for ViT-B-16}
\label{table:vit-arch}
\begin{tblr}{
  vline{2} = {-}{},
  hline{1-2,11} = {-}{},
}
\textbf{Configuration} & \textbf{Value} \\
image resolution       & 384 $\times$ 384      \\
patch resolution       & 16 $\times$ 16        \\
learning rate          & 0.001          \\
weight decay           & 0.0001         \\
batch size             & 16             \\
hidden size            & 768            \\
mlp size               & 3072           \\
\#heads                & 12             \\
encoder length         & 12             
\end{tblr}
\end{table}

\subsection{Sparse Regularization Effect}
\label{sre}
We have performed experiments to evaluate the effectiveness of sparse regularization. There are five scenarios for implementing sparse regularization: similarity score, attention weight, weighted value, the output layer, and the input GELU activation function at the MLP layer. The $\lambda$ value used is $\frac{1}{n\_feature}$. Before testing, the vision transformer model will be fine-tuned on the Cifar 10 and Cifar 100 datasets. The Vision Transformer model without using the sparse regularization method will also be fine-tuned with the same data for comparison. The comparison of sparse regularization placement in the Vision Transformer model is shown in Table \ref{sr-cifar10}.

\begin{table}[h]
\centering
\caption{Result on CIFAR-10}
\label{sr-cifar10}
\begin{tblr}{
  row{1} = {c},
  cell{2}{1} = {c},
  cell{2}{2} = {c},
  vline{2-3} = {-}{},
  hline{1-2,8} = {-}{},
}
\textbf{Layer} & \textbf{Sparse Position} & \textbf{Acc} \\
-              & -                        & 98.83        \\
attention      & similarity score         & 98.57        \\
attention      & attention weight         & 98.81        \\
attention      & weighted value           & 98.73        \\
attention      & output                   & 98.33        \\
MLP            & input GELU               & 98.52        
\end{tblr}
\end{table}

The experiment used the CIFAR-10 dataset and the Vision Transformer model, which had been fine-tuned for 20 epochs and applied with sparse regularization. As a result, after calculating the Attention Weight, the model achieved the highest accuracy of 98.81\% when sparse regularization was applied to the Self-Attention layer. On the other hand, the lowest accuracy of the model with sparse regularization was 98.33\% when the sparse regularization was applied after the output layer calculation in the Self-Attention layer. However, despite sparse regularization, the best results from the model still could not outperform the baseline model, which achieved an accuracy of 98.81\%. The comparison of sparse regularization placement in the Vision Transformer model is shown in Table \ref{sr-cifar10}.

\begin{table}[h]
\centering
\caption{Result on CIFAR-100}
\label{sr-cifar100}
\begin{tblr}{
  row{1} = {c},
  cell{2}{1} = {c},
  cell{2}{2} = {c},
  vline{2-3} = {-}{},
  hline{1-2,8} = {-}{},
}
\textbf{Layer} & \textbf{Sparse Position} & \textbf{Acc} \\
-              & -                        & 92.39        \\
attention      & similarity score         & 91.51        \\
attention      & attention weight         & 92.52        \\
attention      & weighted value           & 92.17        \\
attention      & output                   & 92.13        \\
MLP            & input GELU               & 91.73        
\end{tblr}
\end{table}

The second experiment was conducted using the CIFAR-100 dataset and the Vision Transformer model, which had been fine-tuned for 20 epochs and applied with sparse regularization. As a result, the model achieved the highest accuracy of 92.52\% when sparse regularization was applied to the Self-Attention layer after calculating the Attention Weight. On the other hand, the lowest accuracy was 91.73\% when the sparse regularization was applied to the MLP layer before calculating the GELU activation function. The model with sparse regularization achieved an accuracy increase of 0.12\%, outperforming the baseline model. This indicates that data with more classes makes the vision transformer model have many sparse active neurons. The pruning method will work better if it is applied to CIFAR-100 datasets.

\subsection{Pruning Effect}
\label{pr-e}
\begin{figure}[h]
    \centering
    \includegraphics[width=0.80\linewidth]{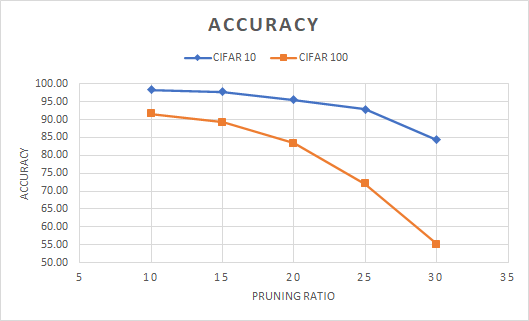}
    \caption{Pruning effect on CIFAR-10 and CIFAR-100}
    \label{fig:pr-e}
\end{figure}

To evaluate the impact of pruning on the performance of the Vision Transformer, we have conducted a comprehensive series of experiments by exploring the effects of varying pruning percentages on both the CIFAR-10 and CIFAR-100 datasets. Through these experiments, we strive to gauge the influence of pruning on accuracy and identify optimal pruning thresholds for the Vision Transformer architecture. We use a range of pruning percentages to achieve this goal, from 10\% to 30\% of the model weight. This systematic approach allowed us to evaluate the pruned Vision Transformer's performance under different weight reduction levels. The pruning was applied globally, ensuring a uniform impact across the entire model. The comparison of accuracy results using the pruning method on CIFAR-10 and CIFAR-100 datasets is shown in Figure \ref{fig:pr-e}.

The experiment was conducted using CIFAR-10 and CIFAR-100 datasets with the ViT-B-16 model, which had been fine-tuned for 20 epochs. After training the model, pruning was performed on all layers with weights. The weights were determined to be pruned using the $l_1$-norm method. As a result, both datasets showed a negative correlation between the decrease in accuracy and the percentage of pruned parameters: the greater the number of pruned parameters, the lower the accuracy. These results underscore the need for careful consideration and optimization during pruning. While pruning offers potential for model compression and computational efficiency, it must be done judiciously to reduce the loss of accuracy. Researchers can navigate this trade-off by leveraging techniques such as the L1-norm method to determine the weights to trim, seeking the ideal balance between model cohesiveness and performance.

\subsection{Effect of Sparse then Prune}
\label{stp}
Based on previous experiments, the best accuracy was achieved by applying sparse regularization after calculating the attention weight on the Self Attention layer.
\begin{figure}[h]
     \centering
     \begin{subfigure}[b]{0.45\textwidth}
         \centering
         \includegraphics[width=\textwidth]{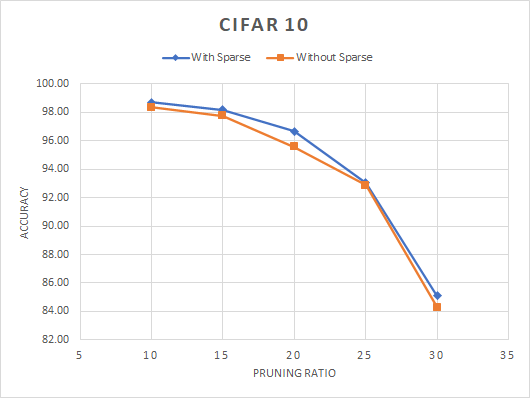}
         \caption{}
         \label{fig:stp-on-c10}
     \end{subfigure}
     \hfill
     \begin{subfigure}[b]{0.45\textwidth}
         \centering
         \includegraphics[width=\textwidth]{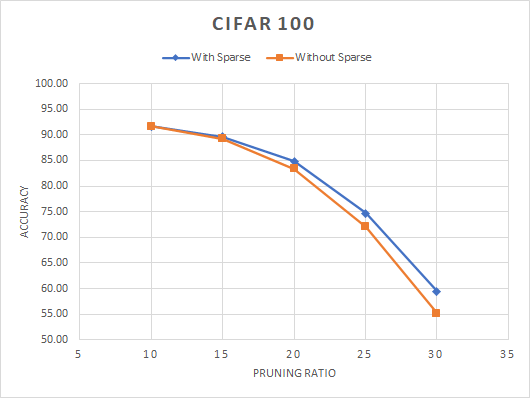}
         \caption{}
         \label{fig:stp-on-c100}
     \end{subfigure}
     \caption{(a) Result on CIFAR-10. (b) Result on CIFAR-100}
\end{figure}
Therefore, in this test, the model was trained with sparse regularization placed after calculating the attention weight on the Self Attention layer before pruning. Similar to the previous test, the pruning percentage used ranged from 10\% to 30\% and was applied globally.

Pruning performed on models with sparse regularization produces higher accuracy than those without sparse regularization. This indicates that sparse regularization can effectively distinguish between important and unimportant parameters or weights. Consequently, when unimportant parameters are pruned or deleted, the model with sparse regularization demonstrates better data classification capabilities than the baseline model. Figures \ref{fig:stp-on-c10} and \ref{fig:stp-on-c100} compare the accuracy of pruning with and without sparse regularization on the Vision Transformer model. In the case of CIFAR-10, the average difference in accuracy is 0.568\%, while for CIFAR-100, the average difference in accuracy is 1.764\%.

\subsection{Result on CIFAR-10 and CIFAR-100}
\label{final-result}

All models are pre-trained on ImageNet before being used in downstream tasks with smaller datasets. A comparison of the Transformer-based specification model can be seen in Table \ref{model-spec}. The model with $\downarrow$10\% means the model is pruned with a 10\% pruning ratio. Besides, the model with $\uparrow$384 is fine-tuned on a bigger resolution, which is 384$\times$384 pixel.

\begin{table}[h]
\centering
\caption{Model Specification}
\label{model-spec}
\begin{tblr}{
  row{1} = {c},
  cell{2}{2} = {c},
  cell{2}{3} = {c},
  cell{2}{4} = {c},
  cell{2}{5} = {c},
  cell{3}{2} = {c},
  cell{3}{3} = {c},
  cell{3}{4} = {c},
  cell{3}{5} = {c},
  cell{4}{2} = {c},
  cell{4}{3} = {c},
  cell{4}{4} = {c},
  cell{4}{5} = {c},
  cell{5}{2} = {c},
  cell{5}{3} = {c},
  cell{5}{4} = {c},
  cell{5}{5} = {c},
  cell{6}{2} = {c},
  cell{6}{3} = {c},
  cell{6}{4} = {c},
  cell{6}{5} = {c},
  cell{7}{2} = {c},
  cell{7}{3} = {c},
  cell{7}{4} = {c},
  cell{7}{5} = {c},
  cell{8}{2} = {c},
  cell{8}{3} = {c},
  cell{8}{4} = {c},
  cell{8}{5} = {c},
  cell{9}{2} = {c},
  cell{9}{3} = {c},
  cell{9}{4} = {c},
  cell{9}{5} = {c},
  cell{10}{2} = {c},
  cell{10}{3} = {c},
  cell{10}{4} = {c},
  cell{10}{5} = {c},
  cell{11}{2} = {c},
  cell{11}{3} = {c},
  cell{11}{4} = {c},
  cell{11}{5} = {c},
  cell{12}{2} = {c},
  cell{12}{3} = {c},
  cell{12}{4} = {c},
  cell{12}{5} = {c},
  vline{2-5} = {-}{},
  hline{1-2,12} = {-}{},
}
\textbf{Model}      & \textbf{\#params} & {\textbf{Image}\\\textbf{Size}} & \textbf{Hidden size} & \textbf{\#heads} \\
ViT-B/16-Sparse     & 86M               & 384                             & 768                  & 12               \\
ViT-B/16-Sparse$\downarrow$10\% & 77M               & 384                             & 768                  & 12               \\
ViT-B/16-Sparse$\downarrow$15\% & 73M               & 384                             & 768                  & 12               \\
ViT-B/16-Sparse$\downarrow$20\% & 69M               & 384                             & 768                  & 12               \\
ViT-B/16-Sparse$\downarrow$25\% & 64M               & 384                             & 768                  & 12               \\
ViT-B/16-Sparse$\downarrow$30\% & 60M               & 384                             & 768                  & 12               \\
ViT-B/16-SAM        & 87M               & 244                             & 768                  & 12               \\
DeiT-B$\uparrow$384          & 86M               & 384                             & 768                  & 12               \\
CeiT-S$\uparrow$384          & 24M               & 384                             & 768                  & 12               \\
TNT-B$\uparrow$384           & 65M               & 384                             & 40+640               & 4+10  \\           
\end{tblr}
\end{table}
%CvT-21           & 32M               & 244                             & -                    & -                

\begin{table}[h]
\centering
\caption{Accuracy comparison on CIFAR-10 and CIFAR-100}
\label{acc-comp}
\begin{tblr}{
  row{1} = {c},
  vline{2-4} = {-}{},
  hline{1-2,12} = {-}{},
}
\textbf{Model}  & \textbf{CIFAR-10} & \textbf{CIFAR-100} & \textbf{Average} \\
ViT-B/16-Sparse & 98.81             & 92.52              & 95.66           \\
ViT-B/16-Sparse$\downarrow$10\% & 98.71             & 91.66              & 95.18           \\
ViT-B/16-Sparse$\downarrow$15\% & 98.16             & 89.62              & 93.89            \\
ViT-B/16-Sparse$\downarrow$20\% & 96.68             & 84.88              & 90.78            \\
ViT-B/16-Sparse$\downarrow$25\% & 93.03             & 74.67              & 83.85            \\
ViT-B/16-Sparse$\downarrow$30\% & 85.09             & 59.47              & 72.28            \\
ViT-B/16-SAM    & 98.2              & 87.6               & 92.9             \\
DeiT-B$\uparrow$384            & 99.1              & 90.8               & 94.95            \\
CeiT-S$\uparrow$384            & 99.1              & 90.8               & 94.95            \\
TNT-B$\uparrow$384             & 99.1              & 91.1               & 95.1       %      \\
%CvT-21             & 99.16             & 92.88              & 96.02            
\end{tblr}
\end{table}

Table \ref{acc-comp} report numerical result on CIFAR-10 and CIFAR-100 as a downstream task. Interestingly, with almost the same parameters as defined in the original paper \cite{DBLP:journals/corr/abs-2010-11929}, Vision Transformer with sparse regularization takes first place on average accuracy. %CvT-21 occupies the first position with a 32M parameter, significantly less than the original Vision Transformer.

\subsection{Result on ImageNet-100}
\label{rs-in-100}
To provide a broader overview of the sparse regularization effect on pruning, this experiment has also been carried out using ImageNet-100, a subset of ImageNet1k. The sampling process was carried out randomly by taking 100 classes from ImageNet1k dataset. As a result, the ViT model with sparse regularization can overcome the original ViT model with 0.12\% higher accuracy. The detail can be seen in Tabel \ref{acc-in-100}.

\begin{table}[h]
\centering
\caption{Accuracy on ImageNet-100}
\label{acc-in-100}
\begin{tblr}{
  cell{1}{1} = {c},
  cell{1}{3} = {c},
  cell{2}{2} = {c},
  vline{2-3} = {-}{},
  hline{1-2,4} = {-}{},
} 

\textbf{Model}  & Sparse Position  & \textbf{Accuracy} \\
ViT-B/16        & -                & 96.8              \\
ViT-B/16-Sparse & Attention Weight & 96.92             
\end{tblr}
\end{table}

\begin{figure}[h]
    \centering
    \includegraphics[width=0.80\linewidth]{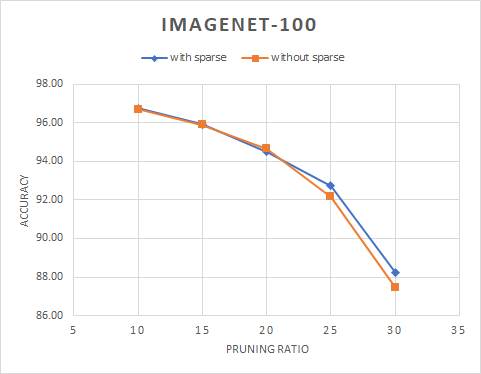}
    \caption{Pruning effect on ImageNet-100}
    \label{fig:sparse-prune-in100}
\end{figure}

Figure \ref{fig:sparse-prune-in100} shows the effect of pruning performed on the model that applies sparse regularization. It can be seen that the result has a similar pattern to the experiment that has been done on CIFAR-10 and CIFAR-100 data. This method produces higher accuracy than pruning performed on models without sparse regularization, with 0.25\% higher accuracy on average. 

\section{Conclusion}
\label{conclusion}
The implementation of sparse regularization will produce the best accuracy if it is placed after the attention weight calculation on the self-attention layer. Pruning on models with sparse regularization produces better accuracy than pruning on models without sparse regularization.

The Vision Transformer model with sparse regularization can improve accuracy by 0.12\% on CIFAR-100 and 0.12\% on ImageNet-100. Meanwhile, on CIFAR-10 data, the model with sparse regularization has yet to outperform the baseline model. While there is a negative correlation between pruning and accuracy, with the accuracy decreasing as the percentage of pruned parameters increases, models with sparse regularization tend to have a slightly higher average accuracy. In particular. On CIFAR 10 data. Pruning on models with sparse regularization achieves an average higher accuracy of 0.568\%.
Similarly, on CIFAR 100 data. The average higher accuracy achieved through pruning on models with sparse regularization is 1.764\% and 0.256\% higher on ImageNet-100.

In summary, our sparse regularization and pruning explorations have revealed a delicate interplay between these techniques and model accuracy. Sparse regularization, when strategically placed, can unlock the true potential of deep learning models, enhancing accuracy in certain contexts. Furthermore, the combination of sparse regularization and pruning presents a compelling approach to mitigating the negative impact of pruning on accuracy.
%% If you have bibdatabase file and want bibtex to generate the
%% bibitems, please use
%%
\bibliographystyle{elsarticle-num}
\bibliography{cas-refs}

%% else use the following coding to input the bibitems directly in the
%% TeX file.

% \begin{thebibliography}{00}

% %% \bibitem{label}
% %% Text of bibliographic item

% \bibitem{}

% \end{thebibliography}
\end{document}